\documentclass[journal]{IEEEtran}

\usepackage{times}
\usepackage{epsfig}
\usepackage{graphicx}
\usepackage{amsmath}
\usepackage{amssymb}
\usepackage{bm}
\usepackage{graphicx} 
\usepackage{multirow}
\usepackage{algorithm}
\usepackage{algpseudocode}
\usepackage{amsmath}
\usepackage{graphics}
\usepackage{epsfig}
\usepackage{booktabs}
\usepackage{cite}
\usepackage{threeparttable}
\usepackage{xcolor}
\usepackage{colortbl}
\definecolor{mygray}{gray}{0.9}
\definecolor{mypink}{rgb}{.99,.91,.95}
\definecolor{mycyan}{cmyk}{.3,0,0,0}
\usepackage{color}

\newcommand{\tabincell}[2]{\begin{tabular}{@{}#1@{}}#2\end{tabular}}
\usepackage{hyperref}

\begin{document}
	
	\title{SASL: Saliency-Adaptive Sparsity Learning for Neural Network Acceleration}
	
\author{Jun~Shi, Jianfeng~Xu, Kazuyuki~Tasaka,
	Zhibo~Chen,~\IEEEmembership{Senior~Member,~IEEE}
	\thanks{Jun Shi and Zhibo Chen are with the CAS Key Laboratory of Technology in Geo-spatial Information Processing and Application System, University of Science and Technology of China, Hefei 230027, China. Jianfeng Xu and Kazuyuki Tasaka are with the KDDI Research, Japan (e-mail: shi1995@mail.ustc.edu.cn; ji-xu@kddi-research.jp; ka-tasaka@kddi-research.jp; chenzhibo@ustc.edu.cn). This work was supported in part by NSFC under Grant U1908209, 61632001 and the National Key Research and Development Program of China 2018AAA0101400.}
}
	\maketitle
	
	\begin{abstract}
		Accelerating the inference of CNNs is critical to their deployment in real-world applications. Among all pruning approaches, the methods of implementing a sparsity learning framework have shown effectiveness as they learn and prune the models in an end-to-end data-driven manner. However, these works impose the same sparsity regularization on all filters indiscriminately, which can hardly result in an optimal structure-sparse network. In this paper, we propose a Saliency-Adaptive Sparsity Learning (SASL) approach for further optimization. A novel and effective estimation of each filter, \textit{i.e.}, saliency, is designed, which is measured from two aspects: the importance for prediction performance and the consumed computational resources. During sparsity learning, the regularization strength is adjusted according to the saliency, so our optimized format can better preserve the prediction performance while zeroing out more computation-heavy filters. The calculation for saliency introduces minimum overhead to the training process, which means our SASL is very efficient. During the pruning phase, in order to optimize the proposed data-dependent criterion, a hard sample mining strategy is utilized, which shows higher effectiveness and efficiency.  Extensive experiments demonstrate the superior performance of our method. Notably, on ILSVRC-2012 dataset, our approach can reduce 49.7\% FLOPs of ResNet-50 with very negligible 0.39\% top-1 and 0.05\% top-5 accuracy degradation.
		
	\end{abstract}
	
	
	\section{Introduction}
	Convolutional neural networks (CNNs) have been widely applied in a variety of computer vision applications, including classification \cite{krizhevsky2012imagenet, szegedy2015going, he2016deep}, object detection \cite{dai2016r,ren2015faster} and semantic segmentation \cite{chen2014semantic, long2015fully}, \textit{etc}. Scaling up the size of models has been the main reason for the success of deep learning. For instance, the depth of ImageNet Classification Challenge \cite{russakovsky2015imagenet} winner models, has evolved from 8 of AlexNet \cite{krizhevsky2012imagenet} to over 100 of ResNet \cite{he2016deep}. Empirically, larger networks can exhibit better performances, while they are also known to be heavily over-parameterized \cite{zhang2016understanding}. 
	
	However, large CNNs might be incompatible with their deployment in real-world applications, suffering severely from the massive computational and storage overhead. Therefore, it is really necessary to obtain the compact networks with efficient inference.
	
	Pruning \cite{han2015deep} is a common approach to slim neural networks via removing the redundant weights, filters and layers. Weight pruning can achieve a higher compression ratio, but will lead to unstructured sparsity of CNNs, which makes it hard to leverage the efficient Basic Linear Algebra Subprograms (BLAS) libraries \cite{luo2017thinet}. Therefore, structured sparsity pruning becomes more attractive since it can reduce the model size as well as accelerate the inference.
	
	Among all structured sparsity pruning approaches, sparsity learning (SL) \cite{
			zhou2016less,wen2016learning, liu2017learning,wang2017novel,louizos2017learning,shi2018feature, ding2018auto,huang2018data, xie2019learning, koneru2019sparse,yun2019trimming,li2019oicsr, lin2019toward, wang2019structured,lemaire2019structured, ayinde2019regularizing, lin2019towards,yang2019deephoyer,gordon2018morphnet
		}, or called sparsity regularization, is a popular and powerful direction these days. These works introduce sparsity regularization on structures during the training phase. Training with structured regularization can transfer significant features to a small number of filters and automatically
	obtains a structure-sparse model \cite{wen2016learning}. However, in these SL approaches, the sparsity regularization imposed on all filters is the same, without the consideration of specified characteristics of different parts of the models. Theoretically and empirically, two main results can be incurred from such indiscriminate regularization. First, the performance of important filters for prediction is also influenced by the equal regularization, so the final prediction precision after SL may drop by a large margin. And sometimes it is difficult to restore the performance after pruning. Second, different filters consume different computational resources. According to our analysis and experimental observations, traditional SL tends to zero out lots of light filters while retaining the computation-heavy filters. Therefore, we can hardly obtain the optimal structure-sparse networks using the indiscriminate SL.
	
	In this paper, we propose a novel SL form, namely Saliency-Adaptive Sparsity Learning (SASL), to learn better compact neural networks. The saliency of a convolutional filter is considered from two aspects: 1) the importance for prediction performance, which is defined as the change in the loss function induced by removing this filter from the neural network. 2) the consumed resources, especially the computational costs. During SL, the regularization for every filter will be adjusted adaptively according to the saliency of filters, and the calculation for saliency will only lead to very minimum overhead. In the pruning phase, saliency is also proven to be a better criterion than other methods, and the proposed hard sample mining strategy can further improve the effectiveness and efficiency of saliency. In brief, the main contributions of this paper are summarized as follows.
	
	\begin{itemize}
		\item We propose a novel form of sparsity learning, the Saliency-Adaptive Sparsity Learning (SASL). Compared with traditional SL, our optimized format can better preserve the performance of models and reduce more computation for inference without too much overhead.
		\item  In the pruning stage, we observe and analyze that saliency is a better criterion than previous methods. Since saliency is data-dependent, a hard sample mining strategy is proposed to further enhance the effectiveness and efficiency of it.
		\item Extensive experiments on two benchmark datasets for various CNN architectures demonstrate the effectiveness and efficiency of the proposed approach. Typically, on ILSVRC-2012 dataset, 49.7\% FLOPs of ResNet-50 are reduced with only 0.05\% top-5 accuracy degradation, which significantly outperforms other state-of-the-art methods. 
	\end{itemize}
	
	The remainder of this paper is organized as follows. we review works related to network pruning and sparsity learning in Section \ref{section 2}. Section \ref{section 3} introduces the motivation of this paper and the details of the proposed SASL approach. The experimental results and corresponding analyses are presented in Section \ref{section 4}. Finally, we conclude our work in Section \ref{section 5}. The code of SASL will be shared on our website: \url{http://staff.ustc.edu.cn/~chenzhibo/resources.html}  for public research. 	
	
	\section{Related Works}
	\label{section 2}
	Network pruning has been a long-studied project ever since the very early stage of neural network. In this section, we review the significant development of network pruning.

	\textbf{Weight Pruning.}
	In the 1990s, Optimal Brain Damage \cite{lecun1990optimal} and Optimal Brain Surgeon \cite{hassibi1993second} were proposed, in which, unimportant weights were removed based on the Hessian of the loss function. In recent work, Han \textit{et al.} \cite{han2015deep} brought back this idea by pruning the weights of which the absolute values are smaller than a pre-set threshold. And Molchanov \textit{et al.} \cite{molchanov2017variational} proposed Variational Dropout to prune redundant weights. Moreover, \cite{baker2016designing} formulated weight pruning into an optimization problem by finding the weights to minimize the loss while satisfying the pruning cost condition.  Kang \cite{kang2019accelerator} proposed a weight pruning scheme considering the accelerator. Frankel \textit{et al.} \cite{frankle2018lottery} developed the lottery ticket theory, based on which, only using the weights of \textit{winning tickets} can present comparable or even better performance than the original model. But finding the winning tickets initialization is complex and computational expensive. To solve this, Morcos \textit{et al.} \cite{morcos2019one} proposed a generalization method which allows reusing the same wining tickets across various datasets. And Ding \textit{et al.} \cite{ding2019global} focused on the optimizer and designed a novel momentum-SGD, which shows superior capability to find better winning tickets. However, the nature of unstructured-sparsity makes it only yield effective compression but cannot speed up the inference without dedicated hardware/library.
	
	\textbf{Structured Pruning.}
	Therefore, much attention has been focused on structured pruning to accelerate the inference of neural networks. Filter pruning, or called channel pruning, is the most common and flexible way of structured pruning, since filter is the smallest unit of structure.
	
	Many heuristic methods are proposed to prune filters based on the handcrafted features. For example, based on the \textit{smaller-norm-less-important} belief, Li \textit{et al.} \cite{li2016pruning} proposed pruning filters according to the filter weight norm. Average percentage of zeros (APoZ) in the output was used in \cite{hu2016network} to measure the importance of filters. He \textit{et al.} \cite{he2017channel} pruned filters to minimize the feature reconstruction error of the next layer. Similarly, \cite{luo2017thinet} pruned redundant filters via estimating the statistics of its next layer. Yu \textit{et al.} \cite{yu2018nisp} implemented feature ranking to obtain importance score and propagated it throughout the network to find the pruned filters.  And Chin \textit{et al.} \cite{chin2018layer} considered pruning as a ranking problem and then compensated the layer-wise approximation error to improve the performance of previous heuristic metrics. \cite{he2018amc} used reinforcement learning to decide the pruned ratio of each layer. He \textit{et al.} \cite{he2018soft} proposed the concept of soft pruning, which allows pruned filters to recover during training. Furthermore, Huang \textit{et al.} \cite{huang2018learning} trained pruning agents to remove structures in a data-driven way. \cite{molchanov2019importance} used Taylor-expansion to estimate the importance of each filter and then iteratively pruned the least important filters. He \textit{et al.} \cite{he2019filter} proposed pruning the redundant filters via geometric median. Zhao \textit{et al.} \cite{zhao2019variational} proposed a variational Bayesian pruning scheme based on the distribution of channel saliency. Liu \textit{et al.} \cite{liu2019metapruning} proposed the concept of MetaPruning, which combined meta learning with evolutionary algorithm to provide an efficient automatic channel pruning approach. These methods directly pruned filters of the unsparsified models, which would erroneously abandon useful features and result in huge accuracy degradation.
	
	\textbf{Sparsity Learning.}
	So recent approaches \cite{zhou2016less,wen2016learning, liu2017learning,wang2017novel,louizos2017learning,shi2018feature, ding2018auto,huang2018data, xie2019learning, koneru2019sparse,yun2019trimming,li2019oicsr, lin2019toward, wang2019structured,lemaire2019structured, ayinde2019regularizing, lin2019towards,gordon2018morphnet,yang2019deephoyer} have adopted sparsity learning to introduce structured sparsity in the training phase. Zhang \textit{et al.} \cite{zhou2016less} incorporated sparsity constraints into the loss function to reduce the number of filters. Similarly, Wen \textit{et al.} \cite{wen2016learning} utilized Group Lasso to automatically obtain structured sparsity during training.  Moreover, Liu \textit{et al.} \cite{liu2017learning} proposed network slimming to apply $\ell_1$-norm on the channel scaling factors, which can reuse the $\gamma$ parameters in Batch-Normalization layers so there is little training overhead. After SL, filters with small scaling factors will be pruned. And \cite{huang2018data} extended this idea, which utilized the scaling factors for coarser structures apart from filters, such as layers and branches. Lin \textit{et al.} \cite{lin2019towards} further used generative adversarial learning to pick the structures for pruning. Morphnet \cite{gordon2018morphnet} proposed iteratively shrinking and expanding a network to automate the neural network design. In the shrinking procedure, the satisfying regularization is imposed on activations to distinguish the necessary filters. In \cite{lemaire2019structured}, a neuron-budget-aware sparsity learning method is proposed to avoid trial-and-error. Compared with directly pruning, these methods obtain structure-sparse neural networks in the training stage. As a consequence, redundant filters could be removed with less accuracy decline. However, the above methods impose the same sparsity regularization on all filters indiscriminately, which is the critical problem to be solved in this paper. Most recently, several works also recognized this problem. Yang \textit{et al.} \cite{yang2019deephoyer} proposed the method of DeepHoyer, which made use of the tool of Hoyer to replace the $\ell_1$ regularizer and achieved high performance.  Yun \textit{et al.} \cite{yun2019trimming} tried to tackle this with the proposed approach of Trimmed- $\ell_1$ regularizer, which allows  the filters with largest norm  penalty-free from regularization. But this optimization method is still quite simple, and as discussed later, Trimmed-$\ell_1$ regularizer can be seen as a special case of our work.

	\section{Proposed Approach}
	\label{section 3}
	\subsection{Motivation}
	We start by showing and analyzing the flaws of previous indiscriminate sparsity learning. Typically, we can formulate the optimization objective of sparsity learning as:
	\begin{equation}
		\label{eq1}
		Loss = f(\mathcal{W}, \mathcal{\mathcal{D}}) + \lambda \cdot \sum_{i=1}^{S}R(w_i).
	\end{equation}
	Here $\mathcal{W}$ is the overall trainable weights of current CNN, while $\mathcal{D} = {(x_i, y_i)}_{i=1}^N$ is the used training set. $f(\mathcal{W}, \mathcal{D})$ is the original loss function of the task, such as cross entropy. And $R(w_i)$ is the sparsity regularization for filter $w_i$. Generally, the form of $R(\cdot)$ can be Group Lasso or $\ell_1$-norm penalty on the structure-corresponding scaling factors, which is widely used to achieve sparsity. $\lambda$ is the hyper-parameter to balance the original loss and the sparsity regularization. Note that all the filters $w$ share the same parameter $\lambda$ indiscriminately.
	
	This is a coarse format of sparsity learning that may lead to critical problems. The sparsity is achieved without guidance, and we will show the drawbacks from two aspects in the following sections \ref{section3.1.1} and \ref{section3.1.2}.
	Actually, we can integrate some prior information into sparsity learning, thus better structure-sparse neural networks can be obtained. 
	\subsubsection{Importance of Filters}
	\label{section3.1.1}
\begin{figure}[h]
		\centerline{\includegraphics[scale=0.77]{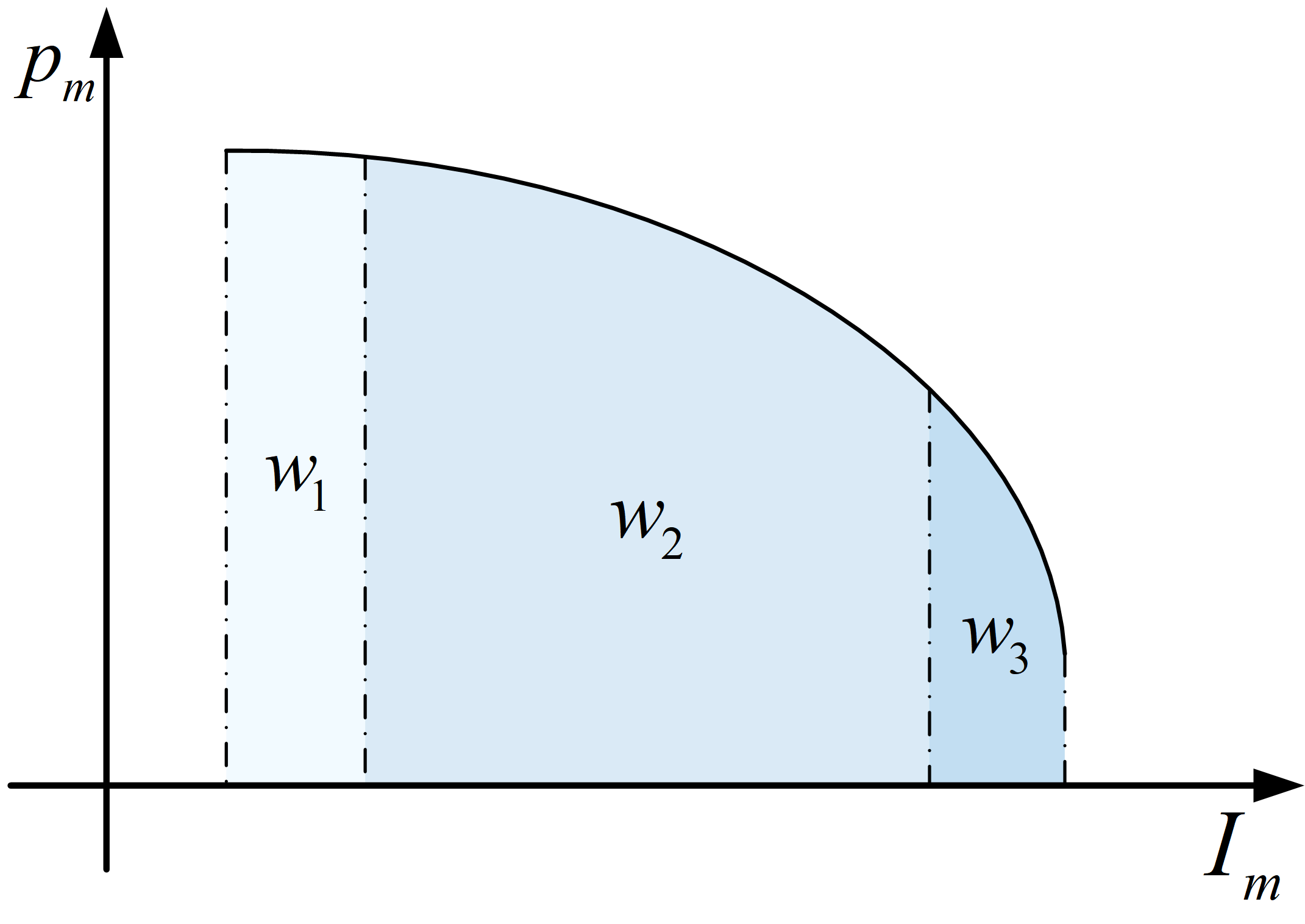}}	
		\vspace{-0.cm}
		\caption{Simplified relationship between importance and pruning possibility of filter $m$. Filters that are important are not likely to be pruned. Deep color denotes high importance. Best viewed in color.}
		\label{figure1}
		\vspace{-0.cm}
	\end{figure}

	Different filters in a neural network are of different importance. The same as \cite{molchanov2019importance}, in this paper we define the importance $\mathcal{I}_m$ of filter $w_m$ as the
	error induced by the removal of it. Under an \textit{i.i.d.} assumption,
	this error can be measured as the squared difference of prediction losses with and without  filter $w_m$: 
	\begin{equation}
		\label{eq2}
		\mathcal{I}_m = (E(\mathcal{D},\mathcal{W}) -E(\mathcal{D},\mathcal{W}|w_m=0))^2.
	\end{equation}
	For a given model $\mathcal{M}$, we denote the corresponding optimal pruned model as $\mathcal{M}'$. And in the transition from $\mathcal{M}$ to $\mathcal{M}'$, the probability of filter $w_m$ to be pruned is denoted as $p_m$. Intuitively, the less important a filter is, the more likely it is to be pruned. So we can assume that the relationship between $\mathcal{I}_m$ and $p_m$ conforms to an inverse correlation function, as Fig. \ref{figure1} shows.
	
	Here we simply divide all filters of a model into three categories. For the least important filters, pruning them directly is not risky. On the contrary, the most significant filters are essential for the prediction precision, and any impact on them will incur a performance decline of the model. The role of sparsity learning is to help identify the filters in the middle part, \textit{i.e.}, $w_2$ in Fig. \ref{figure1}. 
	
	One of the critical issues of indiscriminate sparsity learning is that the importance difference of filters is disregarded. Regularization on the important filters can lead to massive accuracy drop and sometimes this drop cannot be recovered. Moreover, regularization effects on $w_3$ deteriorate the representational capacity of current model, so the sparsity learning would fail to maximally identify the redundant filters.
	\subsubsection{Computational Resources}
	\label{section3.1.2}
	Most of the previous approaches equate the goal of network pruning, \textit{i.e.}, reducing more consumed resources, with removing more structures, such as convolutional filters. However, without the consideration of consumed computational resources (or memory footprint) of different filters, the directions of these two statements are different, and sometimes the gap cannot be ignored.
	Specifically, convolutional filters of different layers in one model would cost different resources. Basically, it depends on the following three factors: 1. input feature map size; 2. number of input feature maps; 3. filter size. Without guidance, traditional sparsity learning such as $\ell_1$-norm scheme can only zero out more filters rather than reducing more computational complexity.
	
	\begin{figure}[h]
		\vspace{-0.cm}
		\centerline{\includegraphics[scale=0.205]{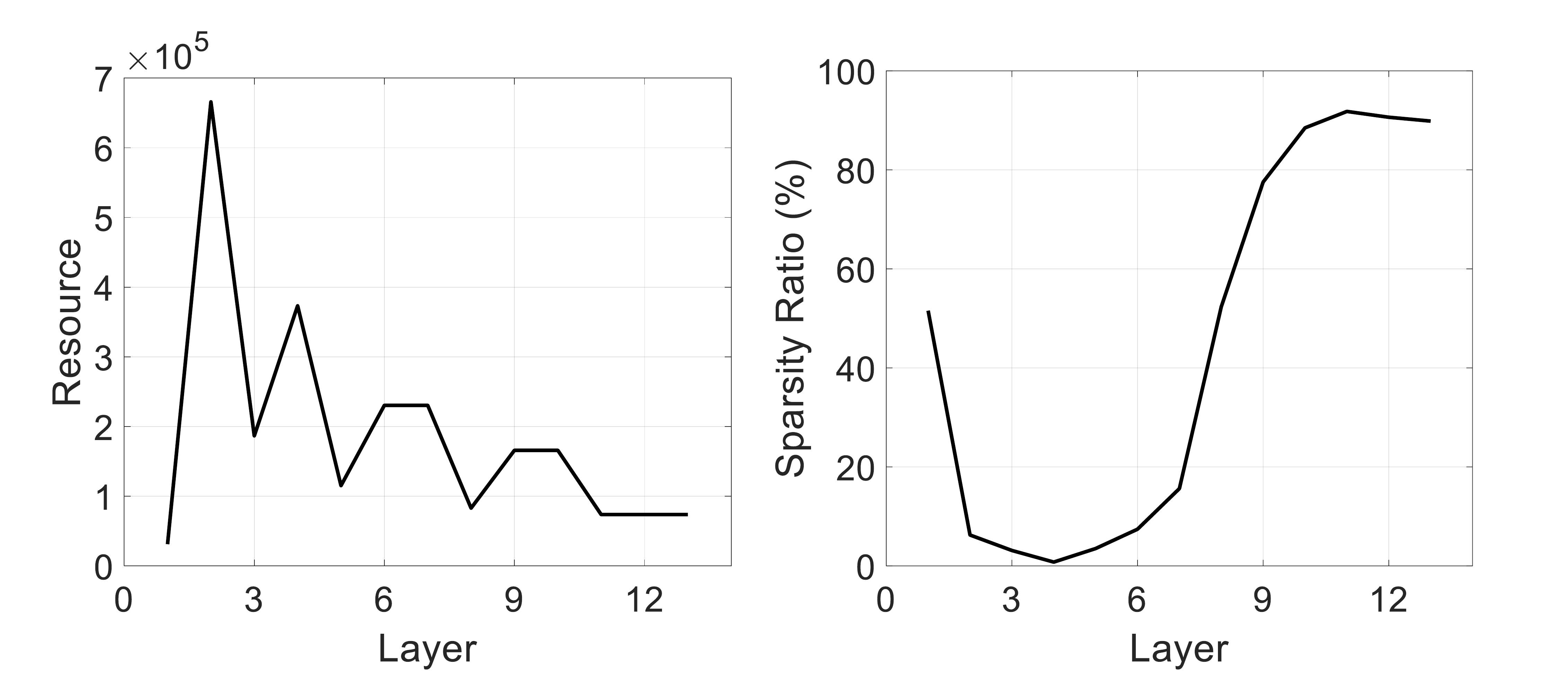}}	
		\caption{Normalized consumed resources and sparsity ratio of filters in different layers of VGG-Net. This strong inconsistency arises from the indiscriminate sparsity learning.}
		\label{figure2}
	\end{figure}
	
	Here we conduct an experiment to show the inconsistency of the indiscriminate sparsity learning. We implement network slimming \cite{liu2017learning} for VGG-Net on CIFAR-10 dataset \cite{krizhevsky2009learning}. When planning to prune 70\% filters, we record the distribution of filters to be pruned which possess the smallest scaling factors. All these scaling factors are less than $10^{-2}$, so we can assume that these filters have been already sparsified. Then we display the normalized computational complexity and sparsity ratio of filters in all layers in Fig. \ref{figure2}.
	
	We can observe that traditional sparsity learning algorithm tends to zero out more light filters, while most of the computation-heavy filters, such as filters in layer $2\sim6$, are retained. Obviously, indiscriminate sparsity learning cannot obtain the optimal structure-sparse networks in terms of complexity reduction.
	\subsection{Saliency Estimation}
	Therefore, in order to solve the critical problems incurred from traditional sparsity learning, we propose imposing adaptive regularization on different filters discriminatively. A new attribute: \textit{saliency}, is introduced to estimate filters and then guide the regularization distribution, which is considered from two aspects: the importance for final prediction and the consumed computational resources.
	
	First we need to estimate the importance $\mathcal{I}$ of all filters during sparsity learning. The most precise evaluation for importance is as Eq. (\ref{eq2}) shows. However, this is extremely computationally expensive, since it requires evaluating all kinds of versions of the neural network, one for each pruned filter. A method to avoid this is to approximate $\mathcal{I}_m$ in the vicinity of $\mathcal{W}$ using second-order Taylor expansion as Optimal Brain Damage \cite{lecun1990optimal} shows:
	\begin{equation}
		\mathcal{I}_m^{(2)}(\mathcal{W}) = (g_m^T\cdot w_m-\frac{1}{2}w_m\cdot H_m^T\cdot \mathcal{W})^2.
	\end{equation}
	Here $g_m = \frac{\partial f(\mathcal{W}, \mathcal{D})}{\partial w_m}$ and $H_{i,j} = \frac{\partial^2 f(\mathcal{W}, \mathcal{D})}{\partial w_i \partial w_j}$ are the gradient and Hessian of filter $w_m$,  respectively. However, computing Hessian matrices sometimes is also computationally expensive, especially for large networks. So we can adopt a more compact approximation, \textit{i.e.}, using the first-order expansion as \cite{molchanov2019importance} does. So the importance can be calculated as:
	\begin{equation}
		\label{eq4}
		\mathcal{I}_m^{(1)}(\mathcal{W}) = (g_m^T\cdot w_m)^2.
	\end{equation}
	
	In this format, calculating the importance will not bring too much computation overhead since $g_m$ is already known from the back-propagation during training.
	
	Then we need to estimate the consumed computational resources $\mathcal{R}$ of different filters. We denote the three influential factors: input feature map size (considering padding pattern and stride), number of input feature maps and filter size as $\mathcal{S}_{fm}$, $\mathcal{N}_{fm}$ and $\mathcal{S}_{f}$, respectively. So the normalized computational resources $\mathcal{R}_m$ of filter $m$ can be calculated as:
	\begin{equation}
		\mathcal{R}_m = \mathcal{S}_{fm}\cdot \mathcal{N}_{fm} \cdot \mathcal{S}_{f}.
	\end{equation}
	Note that the calculation for $\mathcal{N}_{fm}$ should be dynamic in both sparsity learning and pruning, since the number of valid feature maps can be decreased. During sparsity learning, the sparsified filters are excluded, of which the scaling factors are smaller than $10^{-2}$. And in the  pruning phase, the pruned filters will not be counted in.

	Finally, we can calculate the saliency $\mathcal{SAL}_m$ of filter $m$  as:
	\begin{equation}
		\label{eq7}
		\mathcal{SAL}_m = \frac{\mathcal{I}_m}{\mathcal{R}_m} = \frac{(g_m^T\cdot w_m)^2}{\mathcal{S}_{fm}\cdot \mathcal{N}_{fm} \cdot \mathcal{S}_{f}}.
	\end{equation}
	In this definition, saliency can be understood as the average prediction gain with a unit computational cost. We will show in the following sections that saliency is very effective in both sparsity learning and pruning.
	\begin{algorithm}[] 
		\caption{Algorithm Description of SASL} 
		\begin{algorithmic}[1] 
			\Require Training dataset $\mathcal{D}$, initialized model $\mathcal{M}$, number of training epochs $\mathcal{T}$, number of mini-batches $\mathcal{B}$, base sparsity regularization value $\lambda$.
			
			\State Initialize the rank $\mathcal{RA}$ of all filters.
			\For{$t = 1,2,$...$,\mathcal{T}$} 
			\State Distribute sparsity regularization $\vec{\lambda}$ based on $\mathcal{RA}$.
			\For{$b = 1,2,$...$,\mathcal{B}$}
			\State Forward $\mathcal{M}$ to get the prediction loss.
			\State Backward $\mathcal{M}$ to get the gradient $g$.
			\State Record $g$ of all filters.
			\State Optimize $\mathcal{M}$ using $g$ and $\vec{\lambda}$.
			\EndFor
			\State Calculate the saliency $\mathcal{SAL}$ of filters.
			\State Sort filters based on $\mathcal{SAL}$  to update $\mathcal{RA}$.
			\EndFor
			\Ensure Structure-sparse Model $\mathcal{M'}$.
			
		\end{algorithmic} 
		\label{alg.1}
	\end{algorithm}
	
	\subsection{Adaptive Sparsity Regularization}
	
	Based on saliency estimation, we can adaptively set the regularization strength according to this feature of all filters. The indiscriminate format of Eq. (\ref{eq1}) now becomes as:
	\begin{equation}
		Loss = f(\mathcal{W}, \mathcal{D}) +  \sum_{i=1}^{\mathcal{S}}\lambda_i\cdot \mathcal{R}(w_i).
		\label{eq-indis}
	\end{equation}
	Note that the value of regularization factor $\lambda_i$ is now dependent on the filter $w_i$.
	In this paper, we implement this idea based on the scaling factor scheme as network slimming \cite{liu2017learning} does, but it can be easily generalized to all kinds of sparsity learnings. In the scaling factor scheme, for each filter (including the one in convolutional layers and fully-connected layers), a scaling factor is introduced, which is multiplied by the output of the corresponding filter. Then during sparsity learning, the regularization term (\textit{i.e.}, $R(\cdot)$ in Eq. (\ref{eq-indis}) is imposed on these scaling factors. These scaling factors can be seen as the agents to identify the filters. 
	
	Since Batch-Normalization (BN) layer has been widely adopted by most of the modern CNNs, we can reuse the $\gamma$ parameters in BN as the scaling factors. Typically, BN layer performs the following transformation in the network:
	\begin{equation}
		\hat{z} = \frac{z_{in} - \mu_\mathcal{B}}{\sqrt{\sigma^2_\mathcal{B} + \varepsilon}}; \ \ \ 
		z_{out} =  \gamma \hat{z} + \beta.
	\end{equation}
		Here $z_{in}$ and $z_{out}$ are the input feature and output feature of the BN layer, while $\mu_\mathcal{B}$ and $\sigma_\mathcal{B}$ are the values of mean and standard deviation of input features among current batch $\mathcal{B}$. $\gamma$ and $\beta$ are the trainable affine transformation parameters, \textit{i.e.}, scale and shift. Therefore, we can directly leverage the $\gamma$ parameters in BN as the scaling factors since they perform the same function.
		In this way, we would not introduce any additional parameters.
	
	Due to the distinct saliency distributions of different models, directly using the norm of saliency to guide the regularization distribution is not generalized enough. So we propose utilizing saliency to sort all filters, and then based on the ranking of filters, a hierarchy scheme is adopted to adaptively set the sparsity regularization. Typically, for the filters of the most significant class, no regularization will be imposed on, and for the least significant filters, we distribute the strongest regularization penalty. Dedicated design for the hierarchy classification and norm of regularization can lead to better results, but it also requires time-consuming manual working. Therefore, we simply adopt a five-class hierarchy scheme. The filters are classified uniformly according to the saliency score ranking, and the corresponding regularization multiplying factors are set as $\{0\times, 1\times, 2\times, 3\times, 4\times\}$. We will show in the experiments that even such simple design can lead to an excellent result.
	
	Note that the saliency estimation and ranking are all along with the sparsity learning, which means the significant filters at the beginning might be rated as useless during the training. Therefore, the regularization distribution is always dynamic. It can precisely observe the current state and adopt the appropriate action. The detailed algorithm of SASL is summarized in Algorithm \ref{alg.1}.
	
	\subsection{Iterative Pruning with Hard Sample Mining}
	After sparsity learning, an effective criterion is needed to discard filters. Most of the previous sparsity learning approaches, such as \cite{liu2017learning}, prune filters from the energy term, \textit{i.e.}, norm of the scaling factor, due to the "\textit{smaller-norm-less-important}" belief. However, this criterion is not an excellent one since energy cannot fully represent the importance (see in Eq. (\ref{eq4}), importance also takes gradient into consideration) and the consumed resource is not even considered. So we propose using saliency as the criterion for pruning, several advantages of which are listed as follows:
	\begin{enumerate}
		\vspace{-0.cm}
		\item Saliency is effective to estimate filters, from both the importance and resource aspects.
		\vspace{-0.cm}
		\item It is globally consistent throughout the whole network and sensitivity analysis for each layer is not needed.
		\vspace{-0.cm}
		\item This method is able to be applied to any layer in the network, including traditional convolutions, skip connections and fully-connections. It is worth mentioning that pruning skip connections may lead to inconsistency between the feature maps of residual and identity branches. For this situation, we generate the fake all-zero feature maps for the output of each branch if necessary. In this way, the element-wise addition process in skip connections can be done as usual. 
		\vspace{-0.cm}
		\item Although saliency estimates the fine-grained structure, \textit{i.e.}, filter, using saliency to prune can also automatically remove the coarse structures, such as res-block or multi-branch, if the whole structure is thought to be redundant.
		\vspace{-0.cm}
	\end{enumerate}

\begin{figure}[h]
		\centerline{\includegraphics[scale=0.252]{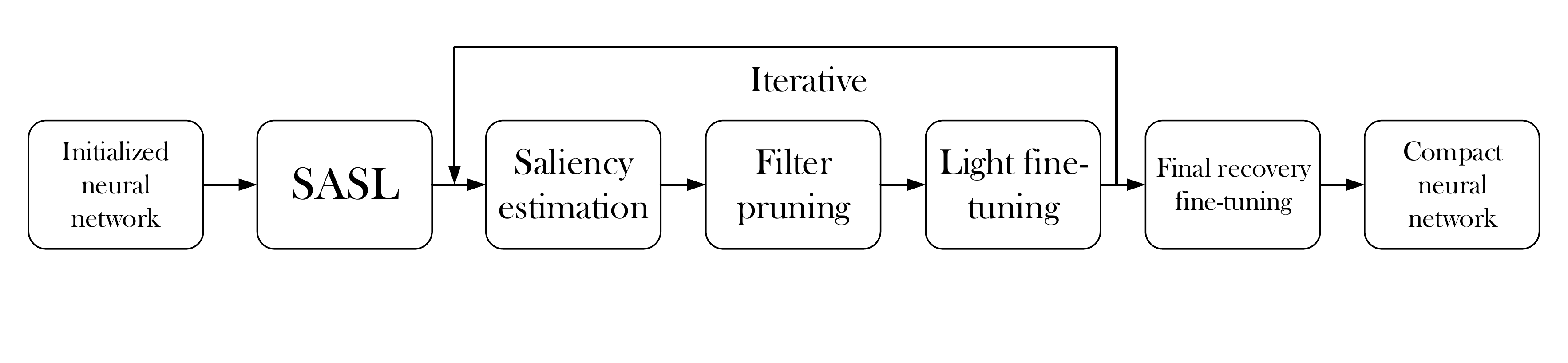}}
		\caption{Proposed iterative pruning framework. We adopt an iterative pruning and fine-tuning strategy, which guarantees precise saliency estimation.}
		\label{figure3}	
	\end{figure}

	The overall proposed pruning procedures are illustrated in Fig. \ref{figure3}. Compared with single pass pruning, an iterative pruning and fine-tuning strategy is adopted to achieve better results, since the estimation for importance and resource is always changing during the pruning process. As a data-dependent metric, saliency is sensitive to the used input data. And one main drawback is the potentially intensive computation because all training data are utilized for saliency estimation during pruning, especially when the training set is huge, such as ILSVRC-2012. And this is much more obvious for our multi-pass pruning scheme.
	However, we do not need to make use of the full training set for saliency estimation. Inspired by OHEM \cite{shrivastava2016training}, in this paper, we propose a hard sample mining approach for optimization. In detail, before pruning, we calculate the training loss of each sample. Then we pick the samples with top 30\% losses, which are defined as the hard samples. And in the pruning phase, we only use the hard samples for saliency estimation. The extra computation of saliency estimation can be dramatically reduced while the pruning effect can be even better than the original scheme which uses the whole training set. We will analyze this in the experiment part. 
	
	\section{Experimental Results and Analyses}
	\label{section 4}
	In this section, we empirically demonstrate the effectiveness of SASL on two benchmark datasets. We implement our method based on the publicly available deep learning framework PyTorch \cite{paszke2017automatic}. We introduce the used datasets and pruned neural networks in \ref{section4.1}, and then present the training details in \ref{section4.2}. In \ref{section4.3} and \ref{section4.4}, we show the experimental results on the two datasets. The complexity analysis of SASL is presented in \ref{complexity}. We implement a series of ablation experiments in \ref{section4.5} to further reveal the superiority of the proposed framework. The running time acceleration analysis is shown in \ref{runningtime}. Finally, we present several discussions in \ref{discussions} to better analyze the special designs.
	\begin{table*}[]
		\centering
		\small
		\renewcommand{\arraystretch}{1.5}
		\caption{Results of Pruning on CIFAR-10 Dataset.}
		\begin{tabular}{llccccc}
			\toprule
			& \multicolumn{1}{c}{Approach} & \tabincell{c}{Baseline \\ Accuracy (\%)} & \tabincell{c}{Pruned \\ Accuracy (\%)} & Accuracy $\downarrow$ (\%)     & Params$\downarrow$ (\%)  & FLOPs$\downarrow$ (\%)  \\\midrule
			\multirow{6}{*}{VGG-Net} & L1 \cite{li2016pruning}                         & 93.25     & 93.40    & -0.15 & 64.0  & 34.2 \\
			& SSS \cite{huang2018data}                         & 93.96      & 93.63    & 0.33\  & 66.7  & 36.3 \\
			& FPGM \cite{he2019filter}                         & 93.58      & 93.54    & 0.04  & -       & 34.2 \\
			& GAL \cite{lin2019towards}                          & 93.96      & 93.42    & 0.54  & 82.2  & 45.2 \\
			& VCP \cite{zhao2019variational}                         & 93.25      & 93.18   & 0.07  & 73.3 & 39.1 \\
			& \textbf{SASL}                         & \textbf{93.69}      & \textbf{93.89}    & \color{red}{\textbf{-0.20}} & \textbf{86.9}  & \color{red}{\textbf{49.5}} \\\midrule
			\multirow{9}{*}{Res-56}  & L1 \cite{li2016pruning}                           & 93.04      & 93.06   & -0.02 & 13.7  & 27.6 \\
			& SFP \cite{he2018soft}                         & 93.59      & 92.26    & 1.33  & -       & 52.6 \\
			& CP \cite{he2017channel}                          & 92.80      & 91.80    & 1.00  & -       & 50.0 \\
			& NISP \cite{yu2018nisp}                         &-              &-            &0.03         &42.6         &43.6       \\
			& FPGM \cite{he2019filter}                         & 93.59      & 93.49    & 0.10  & -       & 52.6 \\
			& GAL \cite{lin2019towards}                          & 93.26      & 93.38    & -0.12 & 11.8  & 37.6 \\
			& VCP \cite{zhao2019variational}                         & 93.04      & 92.26    & 0.78  & 20.5 & 20.3 \\
			& \textbf{SASL}                         &\textbf{93.63}              & \textbf{93.88}    &\color{red}{\textbf{-0.25}}         & \textbf{18.9} & \textbf{35.9} \\
			& \textbf{SASL*}                        &\textbf{93.63}              & \textbf{93.58}    &\textbf{0.05}        & \textbf{36.6} & \color{red}{\textbf{57.1}} \\\midrule
			\multirow{8}{*}{Res-110} & L1 \cite{li2016pruning}                           & 93.53      & 93.30\    & 0.23  & 32.4  & 38.6 \\
			& SFP \cite{he2018soft}                         & 93.68      & 93.38    & 0.30  & -       & 40.8 \\
			& NISP \cite{yu2018nisp}                         &  -            &  -          &  0.18      &43.8         &43.3        \\
			& FPGM \cite{he2019filter}                         & 93.68      & 93.74    & -0.06 & -       & 52.3 \\
			& GAL \cite{lin2019towards}                          & 93.50      & 93.59    & -0.09 & 4.1   & 18.7 \\
			& VCP \cite{zhao2019variational}                          & 93.21      & 92.96    & 0.25  & 41.3 & 36.4 \\
			& \textbf{SASL}                         &\textbf{93.83}              & \textbf{93.99}    &\color{red}{\textbf{-0.16}}         & \textbf{31.9} & \textbf{51.7} \\
			& \textbf{SASL*}                        &\textbf{93.83}              & \textbf{93.80}    &\textbf{0.03}         & \textbf{54.3} & \color{red}{\textbf{70.2}}\\\bottomrule
		\end{tabular}
		\vspace{0.1cm}
		\begin{tablenotes}
			\item SASL and SASL* are the conservative and aggressive schemes, respectively. Accuracy$\downarrow$ is the prediction performance drop between pruned model and baseline model, the smaller, the better. A negative value of Accuracy$\downarrow$ means performance improvement after pruning.
		\end{tablenotes}
		\label{table1}
		
		\vspace{-0.cm}
	\end{table*}
	\subsection{Datasets and Network Models}
	\label{section4.1}
	\subsubsection{Datasets}
	Two classical classification datasets: CIFAR-10 \cite{krizhevsky2009learning} and ILSVRC-2012 \cite{russakovsky2015imagenet}, are adopted in this paper. CIFAR-10 dataset consists of images with resolution 32$\times$32, which is classified into 10 classes. The training and test sets
	contain 50,000 and 10,000 images, respectively.  A standard data augmentation scheme \cite{huang2016deep,lin2013network}, including shifting and mirroring, is used. All input data
	is normalized with channel means and standard deviations.
	
	As for ILSVRC-2012, it is a huge dataset with 1.2 million
	training images and 50,000 validation images which are drawn from 1,000
	classes. We adopt the same data augmentation scheme as PyTorch official examples \cite{paszke2017automatic} for training.
	In the test stage, we will report the single-center-crop validation error of the model as the prediction performance.
	
	\subsubsection{Network Models}
	On CIFAR-10 dataset, we evaluate our framework
	on two popular network architectures: VGG-Net \cite{simonyan2014very} and ResNet \cite{he2016deep}. VGG-Net is originally
	designed for ILSVRC-2012 classification task. In our experiment, a
	variation of  VGG-Net for CIFAR-10 dataset is taken
	from \cite{li2016pruning}. For ResNet, two ResNets of 56 layers and 110 layers are used. On ILSVRC-2012 dataset, we adopt the deep ResNet-50 for pruning. Batch-Normalization layers are adopted in all models to achieve better performance.

	\subsection{Training Details}
	\label{section4.2}
	\subsubsection{Normal Training}
	In normal training, we train all the CNNs  from scratch as baselines. All the models are trained using the optimizer of stochastic gradient descent (SGD). On CIFAR-10 dataset, we train VGG-Net and ResNet using mini-batch size of 64 for 160 and 240 epochs, respectively. 
	The initial learning rate is set as 0.1, and is divided by 10 at 50\%
	and 75\% of the total number of training epochs. And on ILSVRC-2012 dataset, we train ResNet-50 for 90 epochs, with a batch size of 256. The
	initial learning rate is 0.1, and we divide it by 10 after 30
	and 60 epochs. A weight decay of $10^{-4}$
	and a Nesterov momentum \cite{sutskever2013importance} of 0.9 without
	dampening are used in our experiments to improve the training performance. We also adopt the weight initialization introduced by \cite{he2015delving}.
	\begin{table*}[h]
		\centering
		\small
		\caption{Pruning ResNet-50 on ILSVRC-2012 Dataset.}
		\renewcommand{\arraystretch}{1.5}
		\begin{tabular}{lccccccc}
			\toprule
			Approach & \tabincell{c}{Baseline Top-1\\ Acc. (\%)} & \tabincell{c}{Pruned Top-1\\ Acc. (\%)} & \tabincell{c}{Baseline Top-5\\ Acc. (\%)} & \tabincell{c}{Pruned Top-5\\ Acc. (\%)} & \tabincell{c}{Top-1 \\ Acc. $\downarrow$ (\%)}& \tabincell{c}{Top-5 \\ Acc. $\downarrow$ (\%)} & \tabincell{c}{FLOPs $\downarrow$\\ (\%)}\\\midrule
			SFP \cite{he2018soft}      & 76.15              & 74.61            & 92.87              & 92.06            & 1.54      & 0.81      & 41.8  \\
			CP \cite{he2017channel}      & -                  & -                & 92.20              & 90.80            & -         & 1.40      & 50.0  \\
			GDP \cite{lin2018accelerating}      & 75.13              & 71.89            & 92.30              & 90.71            & 3.24      & 1.59      & 51.3 \\
			DCP \cite{zhuang2018discrimination}     & 76.01              & 74.95            & 92.93              & 92.32            & 1.06      & 0.61      & 55.8  \\
			ThiNet \cite{luo2017thinet}   & 72.88              & 71.01            & 91.14              & 90.02            & 1.87      & 1.12      & 55.8  \\
			SSS \cite{huang2018data}      & 76.12              & 74.18            & 92.86              & 91.91            & 1.94      & 0.95      & 31.3  \\
			GAL \cite{lin2019towards}      & 76.15              & 71.80            & 92.87              & 90.82            & 4.35      & 2.05      & 55.0  \\
			Taylor-FO \cite{molchanov2019importance} &76.18  &74.50  & -     & -      &1.68     & -    & 45.0
			\\
			FPGM \cite{he2019filter}     & 76.15              & 75.59            & 92.87              & 92.63            & 0.56      & 0.24      & 42.2  \\
			C-SGD \cite{ding2019centripetal}    & 75.33              & 74.93            & 92.56              & 92.27            & 0.40      & 0.29      & 46.2  \\
			HRank \cite{lin2020hrank}  &76.15  &74.98  &92.87  &92.33  &1.17  &0.54  &43.8\\
			\textbf{SASL}     & \textbf{76.15}              & \textbf{75.76}            & \textbf{92.87}              & \textbf{92.82}            & \color{red}{\textbf{0.39}}      & \color{red}{\textbf{0.05}}      & \textbf{49.7}  \\
			\textbf{SASL*}    & \textbf{76.15}              & \textbf{75.15}            & \textbf{92.87}              & \textbf{92.47}            & \textbf{1.00}      & \textbf{0.40}      & \color{red}{\textbf{56.1}} 
			\\\bottomrule
		\end{tabular}
		\vspace{0.0cm}
		\begin{tablenotes}
		\item SASL and SASL* are the conservative and aggressive schemes, respectively. Acc. $\downarrow$ is the prediction performance drop between pruned model and baseline model, the smaller, the better.
		\end{tablenotes}
		\label{table2}
	\end{table*}
	
	\subsubsection{Sparsity Learning and Pruning}
	Although our framework can adaptively distribute the sparsity regularization, a base regularization value $\lambda$ should be determined in advance, which can control a trade-off between prediction performance and structure sparsity. Empirically, we use relative larger $\lambda$ for the simple VGG-Net ($10^{-4}$), while smaller $\lambda$ for the complicated ResNet ($3\times10^{-5}$).
	Other settings are the same as normal training.
	
In our experiments, the pruning procedure is achieved via building a new compact model and then copying the retained weights from the original model. And saliency is used as the metric to help discard the specified filters from the structure-sparse models. Suppose the number of filters to be discarded is $\mathcal{N}$. In every iteration, we will prune 5\%* $\mathcal{N}$ filters with the least saliency scores. Therefore, 20 pruning iterations are needed. During experiments, we find that the saliency ranking of filters does not change obviously in the early pruning stage. So we also develop a fast pruning scheme, in which we prune 20\%*$\mathcal{N}$ filters at the beginning 3 iterations. In this way, we only need 11 iterations and the final performance is almost the same.
	\subsubsection{Fine-Tuning}
	Using pruning, we can obtain more compact models. Then we need to fine-tune them to restore the performance. The learning rate of fine-tuning for all models is set as $10^{-4}$. On CIFAR-10 datasets, we fine-tune the pruned models for 20 epochs, while on ILSVRC-2012 dataset,
	we only fine-tune the pruned ResNet for 10 epochs.
	
	\subsection{Results on CIFAR-10}
	\label{section4.3}
	For the CIFAR-10 dataset, we test our SASL on VGG-Net, ResNet-56 and ResNet-110.
	As shown in TABLE \ref{table1}, our SASL outperforms other state-of-the-art methods in all three networks. For VGG-Net, SASL reduces 49.5\% FLOPs with even 0.2\% accuracy improvement, while previous works \cite{li2016pruning, huang2018data, he2019filter, lin2019towards, zhao2019variational} are worse in both two aspects. For example, GAL \cite{lin2019towards} only prunes 45.2\% FLOPs and incurs 0.54\% accuracy degradation.
	
	For ResNet-56 and ResNet-110, we prune filters of different ratios to achieve different trade-offs between accuracy and complexity. In TABLE \ref{table1}, SASL means the conservative scheme that tries to preserve the accuracy, while SASL* denotes the aggressive scheme. During sparsity learning, the aggressive scheme slightly increases the base regularization value, and prunes relatively more filters in the pruning stage. Comparing with other works, we can find that our framework also achieves state-of-the-art performance for ResNet. For pruning ResNet-56, SASL* reduces more FLOPs than FPGM \cite{he2019filter} (57.1\% \textit{v.s.} 52.6\%) and better preserves the accuracy (degradation: 0.05\% \textit{v.s.} 0.10\%). On ResNet-110, SASL achieves a higher FLOPs reduction (51.7\% \textit{v.s.} 36.4\%) with 0.16\% accuracy increase, while VCP \cite{zhao2019variational} harms the performance of prediction. These results demonstrate the effectiveness of SASL, which strongly aligns with our previous analyses.
	\subsection{Results on ILVSRC-2012}
	\label{section4.4}
	SASL is also evaluated on ILSVRC-2012 dataset for pruning ResNet-50. Similarly, we adopt both the conservative and aggressive schemes. TABLE \ref{table2} shows the superior performance of SASL. Under various pruned FLOPs ratios, our approach consistently achieves state-of-the-art performance when compared with other methods \cite{he2018soft,he2017channel,lin2018accelerating,zhuang2018discrimination,luo2017thinet,huang2018data,lin2019towards,he2019filter,ding2019centripetal,molchanov2019importance,lin2020hrank}. To be specific, the conservative scheme SASL reduces 49.7\% FLOPs with very negligible 0.39\% top-1 and 0.05\% top-5 accuracy degradation, while SSS \cite{huang2018data} incurs huger performance deterioration (1.94\% top-1 and 0.95\% top-5 accuracy drops) and only prunes 31.3\% FLOPs. Our SASL* also performs well, which reduces more FLOPs than DCP \cite{zhuang2018discrimination} (56.1\% \textit{v.s.} 55.8\%) with better performance maintaining. Compared with previous methods, SASL estimates the saliency of different filters and intelligently distributes the regularization to obtain better structure-sparse networks, which is the main cause of its superior performance.
	
	\begin{table}[!h]
		\centering
		\small
		\caption{Computational Complexity Analysis for One Epoch.}
		\renewcommand{\arraystretch}{1.5}
		\begin{tabular}{llll}
			\toprule
			Action    & Complexity & Frequency & FLOPs   \\ \midrule
			SE        & $O(n)$       & 782       & 4.7$\times10^6$ \\
			SA\&R     & $O(nlogn)$   & 1         & $\leq$ 2.1$\times10^6$ \\
			RD        & $O(n)$       & 1         & 2.0$\times10^3$ \\
			Summation   &   -         &    -       & 6.8$\times10^6$ \\ \midrule
			Inference &     -       &   -        &  6.3$\times10^{12}$       \\ \bottomrule
		\end{tabular}
		\begin{tablenotes}
			\item SE, SA\&R and RD refer to saliency estimation, saliency aggregation \& ranking and regularization distribution, respectively. Inference complexity considers all 50,000 samples in CIFAR-10.
		\end{tablenotes}
		\label{tab-complex}
	\end{table}
	\subsection{Computational Complexity Analysis of SASL}
	\label{complexity}
	Here we analyze the extra complexity of our SASL in addition to the original training procedure. The introduced computational complexity arises from three parts: 1. saliency estimation for every batch; 2. saliency aggregation and ranking for every epoch; 3. hierarchy regularization distribution for every epoch. Actually, since the necessary variables for saliency estimation can be directly obtained from the training procedure, the computational complexity of these three processes is really negligible. We use ResNet-56 on CIFAR-10 as an example, which possesses 2,032 filters (including nodes in fully-connected layers). The detailed complexity for one training epoch has been shown in TABLE \ref{tab-complex}.  It is worth mentioning that the FLOPs value is approximated, in which the comparing operation is also counted.

	From TABLE \ref{tab-complex}, we can find that the extra computational complexity is really small, far less than 1\% of the inference complexity of ResNet-56, let alone the complexity of other processes of training, such as back-propagation and parameter updating. Therefore, our SASL brings minimal overhead to the overall training procedure.
	\subsection{Ablation Study}
	\label{section4.5}
	In this part, we conduct a series of ablation experiments to validate the effectiveness of proposed schemes. For simplicity and reliability, all the following experiments are conducted on CIFAR-10 for ResNet-56. Without specification, the hyper-parameter setting is the same as stated in \ref{section4.2}.
	\subsubsection{Different Sparsity Regularization}
	First, we analyze the effectiveness of the special saliency-adaptive sparsity learning. For comparison, we run the experiment of traditional indiscriminate sparsity learning as baseline. To better show the insight, we also replace the regularization guider, \textit{i.e.}, saliency, with two of its factors, importance and resource. After all kinds of sparsity learnings, we prune filters of different ratios of the models to get a close complexity reduction. Then we fine-tune all the models and show the  classification accuracy fluctuation in TABLE \ref{table3}. In this table, we can see that SASL works much better than the indiscriminate one, with 0.47\% accuracy improvement. The importance- and resource-guided versions can also improve the performance than baseline, but they are both worse than the integrated version. Saliency gives attention to both aspects so it can better guide the sparsity learning.
	
	\begin{table}[h]
		\small
		\centering
		\renewcommand{\arraystretch}{1.1}
		\vspace{-0.cm}
		\caption{Results of Different Regularizations for Sparsity Learning.}
		\begin{tabular}{lcccc}
			\\\toprule
			& Tradition   & Importance & Resource & Saliency \\\midrule
			\multicolumn{1}{l|}{FLOPs $\downarrow$} & 57.0\% & 57.0\% & 57.1\%   & 57.1\%   \\
			\multicolumn{1}{l|}{ACC $\downarrow$}   & 0.52\% & 0.36\% & 0.33\%   & 0.05\%  
			\\\bottomrule
		\end{tabular}
		\vspace{0.cm}
		\label{table3}
		\vspace{-0.cm}
	\end{table}
	\subsubsection{Hierarchy Scheme Extension}
\begin{figure}[h]
		\vspace{-0.cm}
		\centerline{\includegraphics[scale=0.25]{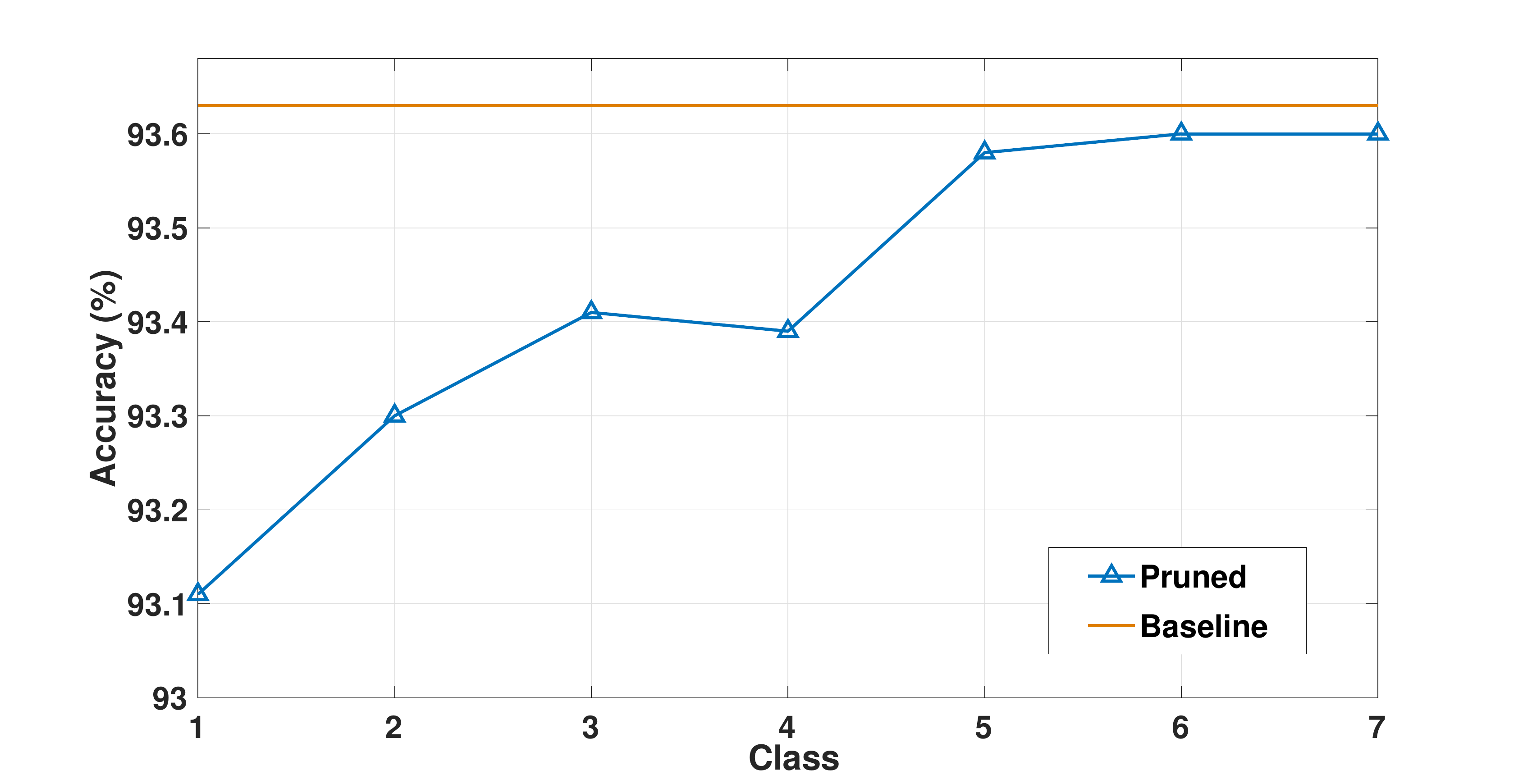}}	
		\caption{Results of different settings for number of classes. With the increase of classes, the final accuracy is improved.}
		\label{figure4}
		\vspace{-0.cm}
	\end{figure}

	Based on the saliency, we classified the filters into five classes and then adaptively impose the regularization. Here we change the hierarchy scheme with different number of classes to explore the influence of this parameter. One-class scheme equals the traditional sparsity learning and five-class scheme is the proposed one.  We also change the base regularization value according to the number of classes so as to impose the same amount of regularization. The regularization multiplier $r_k$ for class number of $k$ is set to satisfy the following equation:
	\begin{equation}
		r_k\cdot \sum_{i=0}^{k-1}i\cdot \frac{1}{k}\cdot r_{base} = \sum_{i=0}^{4}i\cdot \frac{1}{5} \cdot r_{base}.
		\label{eq-class}
	\end{equation}
	Here $r_{base}$ is regularization value for the scheme of five classes. The results after pruning and fine-tuning are shown in Fig. \ref{figure4}. We can find that with the increase of classes, there is a growing in final accuracy, and the increase is slow down when the number of classes is already large. In order to be more generalized for all models and avoid being too complicated, we adopt the scheme of five classes at last. Note that fine-tuning this parameter for the specific neural network may even lead to better results.
	
	We can observe that there is a performance drop at class = 4. Here we present a possible explanation. Generally, for the scheme with more classes, unimportant filters will be imposed on larger regularization, while this is not always the case. To be specific, in our experiment, the base regularization multiplier of class = 4 is a little bit smaller than class = 3 (refer to Eq. (\ref{eq-class})). As a consequence, there is an overlap in the regularization strength on some filters between these two schemes. In other word, at class = 4, some unimportant filters will be imposed on smaller regularization, and some important filters will be imposed on larger regularization, given that both schemes provide the same amount of regularization. These special cases could lead to performance drop. With the number of classes going up, the overlapping areas are gradually eliminated (since the classification for filters is more fine-grained) and the distribution for regularization is more rational, so the general trend of performance is rising up.
	
	\subsubsection{Saliency as Criterion}
	After sparsity learning, we need to adopt a criterion to discard filters. In this paper, we claim that saliency is also an excellent metric, which takes both the importance and resource term into consideration.  Here we compare saliency with other criteria for pruning to show the superiority. The most common criterion in previous sparsity learning approaches is based on the energy term, \textit{i.e.}, the norm of scaling factors or mean value of filters. We also prune the filters from the aspects of importance and resource. The difference between saliency and resource is that the importance factor in saliency (in Eq. (\ref{eq7})) is replaced with the energy term. TABLE \ref{table4} shows the accuracy results of reducing the same ratio of FLOPs. Not surprisingly, saliency is better than other criteria.
	
	\begin{table}[h]
		\vspace{-0.cm}
		\small
		\centering
		\caption{Results of Different Criteria for Pruning.}
		\renewcommand{\arraystretch}{1.1}
		\begin{tabular}{lcccc}
			\\\toprule
			& Energy   & Importance & Resource & Saliency \\\midrule
			\multicolumn{1}{l|}{FLOPs $\downarrow$} & 57.0\% & 56.9\% & 57.2\%   & 57.1\%   \\
			\multicolumn{1}{l|}{ACC $\downarrow$}   & 0.25\% & 0.22\% & 0.18\%   & 0.05\%  
			\\\bottomrule
		\end{tabular}
		\vspace{0.cm}
		\label{table4}
		\vspace{-0.cm}
	\end{table}
	
	\subsubsection{Input Data for Saliency Estimation}
	The proposed criterion for pruning, \textit{i.e.}, saliency, is data-dependent, which means saliency estimation could be sensitive to the used input data. Directly using all training data would bring huge complexity overhead, especially for the multi-pass pruning scheme. In this paper, we propose a hard sample mining strategy for efficient and effective saliency estimation.  We compare it with using all the training data for saliency estimation to prune filters. Surprisingly, hard sample mining strategy can not only reduce the complexity overhead for the saliency estimation, but also improve the overall performance (93.58\% \textit{v.s.} 93.51\%). We attribute the success to the correlation between hard samples and test sets. Easy samples cannot provide too much information for guidance, and sometimes such information can be deemed as the noise, which would influence the accurate pruning action. Preserving the performance for hard samples can make the model work better on the test set.
	
	\subsubsection{Staircase Regularization Distribution} In our approach, we use a staircase hard multiplier \{0$\times$, 1$\times$, 2$\times$, 3$\times$, 4$\times$\} to adjust the regularization, rather than directly applying saliency to construct the multiplier.  This is due to the following two reasons: 1. the range of saliency scores is quite large; 2. models always differ from each other on the distribution of saliency scores. For example, the saliency scores of filters (aggregated for one epoch) in normally trained ResNet-56 range from 1.46$\times 10^{-10}$ to 3.20$\times 10^{-3}$, while it is from 4.38$\times 10^{-18}$ to 1.47$\times 10^{-2}$ for VGG-Net. As a consequence, it is not convenient to directly apply the value of saliency as the guider. 

Here we show the experiment results of using saliency as the multiplier. For the filters with large saliency scores, they are supposed to be important, so we should decrease the regularization strength. Therefore, the reciprocal of saliency, denoted as $RS$, can be an effective guider. In the first trial, we directly use $RS$ as the multiplier and conduct an experiment. Not surprisingly, the neural network cannot be trained since the overall loss function is overwhelmed by the regularization, considering that the max value of $RS$ is so large. The cross-entropy loss cannot provide any effective guide for the training. Therefore, we multiply $RS$ with $10^{-7}$ (based on a grid search) to alleviate the strength of it and conduct another experiment. In this trial, the final accuracy of the pruned model reaches 92.59\%, lower than our designed staircase scheme (accuracy = 93.58\%). We can see that directly using saliency to guide the regularization makes it inconvenient and time-consuming to control the regularization amount, and also leads to sharp feedback effect on the filters, thus the performance is deteriorated.
	
	\subsection{Running Time Reduction}
	\label{runningtime}
	In addition to the metric of FLOPs reduction, we also measure the running time of pruned models and original models to show the effect of realistic acceleration. TABLE \ref{tab-time} shows the results, including VGG-Net on CIFAR-10 and ResNet-50 on ILSVRC-2012. We can observe that there is a gap between the FLOPs reduction rate and running time reduction rate, which may be due to the limitation of IO delay and buffer switch. Also, the acceleration performance of the plain network, \textit{i.e.}, VGG-net, is much better, considering the fact that ResNet is much more complicated. The skip-connections will occupy more time for data flow and transmission.  Note that the running time acceleration requires no special libraries/hardware.
	
\begin{table}[h]
		\vspace{-0.cm}
		\small
		\centering
		\caption{Comparison on FLOPs and Running Time Reduction.}
		\renewcommand{\arraystretch}{1.5}
	\begin{tabular}{llcc}
		\toprule
		Model                       & Scheme & FLOPs $\downarrow$(\%) & Running Time $\downarrow$(\%) \\ \midrule
		VGG-Net                     & SASL   & 49.5  & 46.2         \\ \midrule
		\multirow{2}{*}{ResNet-50}  & SASL   & 49.7  & 37.8         \\
		& SASL*  & 56.1  & 41.8         \\ \midrule
	\end{tabular}
		\begin{tablenotes}
			\item VGG-Net is on CIFAR-10 dataset, while ResNet-50 is on ILSVRC-2012 dataset.		\end{tablenotes}
	\label{tab-time}
\end{table}
	\subsection{Discussions}
	\label{discussions}
	Based on the experimental results and comparisons with other approaches, here we present several discussions to better analyze our special designs.
	\subsubsection{CE Loss as  Compensation for Regularization}
	As we point out earlier, in traditional indiscriminate sparsity learning, the regularization term, \textit{i.e.}, $R(\cdot)$ in Eq. (\ref{eq1}), is without guidance, while the original objective function $f(\mathcal{W}, \mathcal{D})$, such as cross entropy (CE) loss in classification task, can compensate the regularization effect to some extent. The CE loss’s gradient will weaken the effect of the regularizer if the filter is truly important to the final performance. However, this compensation effect is still weak, due to the non-convex optimization procedure of current deep learning framework from back-propagation. As previous method \cite{liu2017learning} shows, one way to avoid removing important filters is to reduce the regularization strength. The drawback of this method is that it requires a multi-pass "SL-pruning" iteration to obtain enough sparisified filters, which is very inefficient and computationally expensive. In comparison, our design makes use of more prior information to provide a hierarchy scheme, which allows larger regularization. Therefore, efficient single-pass SL while preserving the prediction performance is possible.
	
	In addition, the above guidance from CE loss only considers the importance term. In our work, consumed resources term is also integrated into the estimation metric (saliency), which can better guide the sparsity direction for FLOPs reduction.
	\subsubsection{Performance Improvement after Pruning}
	As we can observe in TABLE \ref{table1}, several pruned models achieves higher prediction precision after SASL. For example, when reducing 35.9\% FLOPs of ResNet-56 on CIFAR-10, we can improve the performance with 0.25\% (from 93.63\% to 93.88\%). We hypothesize this is due to the regularization effect of sparsity learning, which naturally selects important features in intermediate layers of a neural network. This effect can remove the redundancy as well as the noisy information. This phenomenon is obvious on the simple task such as CIFAR-10 classification,  while for pruning networks on the complex ILSVRC-2012 dataset, the performance improvement is not very evident since the redundancy of original models is much smaller.
	
	\subsubsection{Comparison with Trimmed-$\ell_1$}
The most recent work \cite{yun2019trimming} also recognized the critical problem of traditional indiscriminate sparsity learning and proposed the method of Trimmed-$\ell_1$ regularizer, in which, filters with the largest norm will not be imposed with regularization, and \cite{yun2019trimming} only implemented this on the simple MNIST dataset for LeNet-5 \cite{lecun2015lenet}.

Compared with Trimmed-$\ell_1$, our work optimizes sparsity learning and works better in two aspects. First, as pointed out earlier, the main belief \textit{smaller-norm-less-important} of \cite{yun2019trimming} works not very well, so the distinguishing of filters only based on the norm cannot be very precise. In comparison, our proposed metric, \textit{i.e.}, saliency, is integrated with the Taylor-expansion approximated importance and consumed computational resources, which can better represent the significance of filters. Second, \cite{yun2019trimming} only leaves several filters to be penalty-free, which can be viewed as the two-class hierarchy scheme, a special case of our work. As seen in Fig. \ref{figure4},  the simple two-class scheme cannot help search the optimal structure-sparse networks effectively.
	\section{Conclusion}
	\label{section 5}
	Current deep convolutional neural networks are effective with high inference complexity. In this paper, we first analyze the critical problems of previous indiscriminate sparsity learning approach and then propose a novel structured
	regularization form, namely SASL, which can distribute the regularization value for each filter according to saliency adaptively. SASL can better preserve the performance and zero-out more computation-heavy filters. We also propose using saliency as the criterion for pruning. To further improve the effectiveness and efficiency of this data-dependent criterion, we utilize a hard sample mining strategy, which shows better performance and also saves computational overhead. Experiments demonstrate the superiority of SASL over other state-of-the-art methods. In future work, we plan to investigate how to combine SASL with
	other acceleration algorithms that are orthogonal to our scheme, such as matrix decomposition, to obtain better performance.

\ifCLASSOPTIONcaptionsoff
  \newpage
\fi

\bibliographystyle{IEEEtran}
\bibliography{IEEEexample,reference}

\end{document}